\documentclass[journal]{IEEEtran}
\usepackage{graphicx}
\usepackage{amsmath}
\usepackage{amssymb}
\usepackage{cite}
\usepackage{hyperref}
\usepackage{algorithm}
\usepackage{subcaption}
\usepackage{algpseudocode}

\begin{document}

\title{Hybrid Meta-Learning Framework for Anomaly Forecasting in Nonlinear Dynamical Systems via Physics-Inspired Simulation and Deep Ensembles}
\author{A.~Burkan~Bereketoglu,~\IEEEmembership{Student Member,~IEEE}%
\thanks{A. B. Bereketoglu is with the Department of Computer Engineering, Middle East Technical University (METU), Ankara 06800, Turkey (e-mail: burkan.bereketoglu@metu.edu.tr).}}
\maketitle

\begin{abstract}
We propose a hybrid meta-learning framework for forecasting and anomaly detection in nonlinear dynamical systems characterized by nonstationary and stochastic behavior. The approach integrates a physics-inspired simulator that captures nonlinear growth-relaxation dynamics with random perturbations, representative of many complex physical, industrial, and cyber-physical systems. We use CNN-LSTM architectures for spatio-temporal feature extraction, Variational Autoencoders (VAE) for unsupervised anomaly scoring, and Isolation Forests for residual-based outlier detection in addition to a Dual-Stage Attention Recurrent Neural Network (DA-RNN) for one-step forecasting on top of the generated simulation data. To create composite anomaly forecasts, these models are combined using a meta-learner that combines forecasting outputs, reconstruction errors, and residual scores. The hybrid ensemble performs better than standalone models in anomaly localization, generalization, and robustness to nonlinear deviations, according to simulation-based experiments. The framework provides a broad, data-driven approach to early defect identification and predictive monitoring in nonlinear systems, which may be applied to a variety of scenarios where complete physical models might not be accessible.
\end{abstract}


\section{Introduction}

In many different fields of science, industry, and engineering, nonlinear dynamical systems are common. Because of nonlinear growth, saturation effects, feedback loops, and sporadic external shocks, these systems frequently display intricate, nonstationary behaviors. Process industries, smart energy grids, cyber-physical systems, industrial control systems, and critical infrastructure monitoring are real-world examples.

Predictive maintenance, operational safety, and system optimization all benefit from having robust and accurate forecasting and early anomaly detection in their systems. Conventional model-based control and monitoring techniques frequently call for precision physical models and expert-based tuning, which could not translate well to other fault circumstances or operational regimes. Therefore needing an expert at all-times. However, when dealing with data scarcity, nonstationary dynamics, or fault conditions that have not yet been identified, such, expert-based or direct data-driven machine learning (ML) systems might not perform well.

To address these challenges, hybrid approaches that combine physics-based simulation with machine learning have gained increasing attention, even if simulators are not that precise. Physics-informed simulations can generate labeled synthetic datasets by providing system parameters and embedding domain priors that reflect typical nonlinear behaviors. Meanwhile, advanced ML models have demonstrated strong forecasting and anomaly detection capabilities across multiple domains, such as intelligent transportation systems \cite{li2024darnnbus}, electric arc furnace control \cite{godoy2022arc}, water quality prediction \cite{li2019water}, sustainability metrics forecasting \cite{huang2022da-skip}, and real-time streaming anomaly detection \cite{duraj2025lstmcnn}.

In this work, we propose a hybrid anomaly forecasting framework that integrates:

\begin{itemize}
  \item \textbf{A Nonlinear Simulator}, capturing hybrid growth-relaxation dynamics with random perturbations, representing general nonlinear systems;
  \item \textbf{Dual-Stage Attention RNN (DA-RNN)} for one-step ahead forecasting \cite{qin2017dual};
  \item \textbf{CNN-LSTM} networks for spatio-temporal feature extraction \cite{elsayed2021cnn-lstm,duraj2025lstmcnn};
  \item \textbf{Variational Autoencoders (VAE)} for reconstruction-based anomaly scoring \cite{zhao2022dalstmvae};
  \item \textbf{Isolation Forest} for residual outlier detection \cite{ma2024isolation}.
\end{itemize}

Unlike conventional threshold-based anomaly detection, the proposed framework monitors forecast deviation to identify abnormal system evolution. By learning nominal system behavior, the forecasting models serve not only for anomaly localization but also to evaluate the potential impact of candidate control actions before they are executed. Predictive evaluation of the system is enabling to forecast state changes that can lead to undesirable behavior, leading to better operational safety even in the absence of a fully known physical model. 

Simulation-based experiments demonstrate that the hybrid meta-learning ensemble consistently outperforms the individual forecasting (DA-RNN, CNN-LSTM), reconstruction (VAE), and residual-based (Isolation Forest) models in anomaly localization, generalization, and robustness under nonlinear regime shifts. Framework that is proposed is offering a transferable and data-efficient approach for predictive monitoring of nonlinear systems even in the absence of full physical models.

\section{Related Work}

\subsection{Hybrid Modeling of Nonlinear Dynamical Systems}

Modeling non-linear dynamical systems often requires a balance between physics-based models and data-driven learning approaches, for computational efficiency and reliability optimization. Physics-informed simulators enable the generation of synthetic training data by varying operational parameters while embedding domain knowledge, especially in cases where real-world data may be limited or expensive to collect. Recent approaches such as Sparse Identification of Nonlinear Dynamical Systems (SINDy) \cite{brunton2016discovering} and Physics-Informed Neural Networks (PINNs) \cite{raissi2019physics} offer ways to incorporate physical structure directly into learning processes, improving interpretability and generalization when exact governing equations are unknown.

\subsection{Attention-Based Forecasting Models}

Recurrent neural networks (RNNs) have shown strong forecasting capabilities for non-linear and non-stationary time-series systems. The Dual-Stage Attention RNN (DA-RNN) architecture enhances forecasting by learning both input-level(feature-level) and temporal-level attention weights to highlight relevant features and critical time steps. Qin \emph{et al.} \cite{qin2017dual} originally introduced DA-RNN for time-series forecasting, and further extensions have applied attention-based forecasting in domains such as energy load prediction \cite{ozcan2021energy}, sustainability monitoring \cite{huang2022da-skip}, and transportation systems \cite{li2024darnnbus}.

\subsection{CNN-LSTM Hybrid Models for Spatio-Temporal Learning}

CNN-LSTM hybridized models combine local feature extraction through convolutional filters with sequential temporal modeling through LSTM layers. This combination has demonstrated improved predictive accuracy in spatio-temporal anomaly detection tasks. Elsayed \emph{et al.} \cite{elsayed2021cnn-lstm} and Duraj \cite{duraj2025lstmcnn} applied these architectures for real-time anomaly detection in network traffic and streaming data scenarios.

\subsection{Generative Models for Unsupervised Anomaly Scoring}

Variational Autoencoders (VAEs) learn low-dimensional latent representations of standard behavior, allowing anomalies to be marked based on reconstruction errors. Zhao \emph{et al.} \cite{zhao2022dalstmvae} demonstrated high accuracy in key performance indicator (KPI) monitoring using hybrid attention-based LSTM-VAE architectures for time-series anomaly detection.

\subsection{Residual-Based Outlier Detection}

Isolation Forest is a widely used algorithm for residual-based outlier detection, especially in high-dimensional forecasting of residual spaces. When combined with forecast models, Isolation Forests can identify unpredictable or previously unseen deviations. Ma \emph{et al.} \cite{ma2024isolation} provides a comprehensive survey of isolation-based anomaly detection methods.

\subsection{Meta-Learning and Multi-Model Fusion}

Ensemble based Meta-Learners integrate multiple forecasting models, reconstruction models, and residual detectors offering improved robustness under distribution shifts and regime changes. Laptev \emph{et al.} \cite{laptev2015generic} and Malhotra \emph{et al.} \cite{malhotra2016lstmenc} demonstrated the benefits of multi-model ensembles for real-world anomaly detection across sensor networks, process industries, and streaming systems.

\subsection{Cross-Domain Applications of Hybrid Forecasting Models}

Several related forecasting and anomaly detection models related to this study have demonstrated successful applications on various domains involving nonlinear time series and complex systems as such:

\begin{itemize}
  \item \textbf{Public Transportation:} Bus arrival time prediction using DA-RNN in intelligent transportation systems \cite{li2024darnnbus};
  \item \textbf{Industrial Forecasting:} Temperature behavior forecasting in electric arc furnaces \cite{godoy2022arc};
  \item \textbf{Environmental Monitoring:} Water quality prediction with recurrent neural networks and evidence theory \cite{li2019water};
  \item \textbf{Sustainability Metrics:} Periodic time-series forecasting for resource use and environmental sustainability \cite{huang2022da-skip};
  \item \textbf{Load Forecasting:} Dynamic drift-adaptive LSTM models for energy interval load forecasting \cite{bayram2023dalstm};
  \item \textbf{Multi-Sensor Anomaly Detection:} LSTM encoder-decoder frameworks for unsupervised anomaly detection in sensor networks \cite{malhotra2016lstmenc}.
\end{itemize}

These applications of various different hybrid learning frameworks for forecasting and anomaly detection showcases usability of such meta-learners across non-linear and non-stationary systems.

\section{Problem Definition}

Let $X \in \mathbb{R}^{N \times D}$ denotes the multivariate input sequence representing system drivers or exogenous variables (e.g., control inputs, environmental conditions, or auxiliary measurements), and let $Y \in \mathbb{R}^N$ denote the primary response variable of the nonlinear system under monitoring. 

The objective of the estimation in the forecasting schema is to predict next-step $\hat{y}_{t+1}$ based on prior observations over a sliding window of size $T$. The forecasting model $f$ minimizes the mean squared forecast error over the available dataset as:

\begin{equation}
\mathcal{L}_{\mathrm{forecast}} = \frac{1}{N-T} \sum_{t=T}^{N-1} \left( y_{t+1} - \hat{y}_{t+1} \right)^2.
\end{equation}

Robust capture of predictability and deviations, we have formulated an ensemble anomaly scoring mechanism that integrates multiple information sources. The composite anomaly score at each time step $t$ is computed as:

\begin{equation}
A_t = \alpha\,\widehat{R}_t + \beta\,S_t + \gamma\,E_t + \delta\,I_t,
\end{equation}

where:
\begin{itemize}
    \item $\widehat{R}_t$ is the normalized forecast residual, measuring one-step prediction error;
    \item $S_t$ represents the learned attention sparsity from the DA-RNN temporal attention mechanism, highlighting structural deviations in feature relevance;
    \item $E_t$ denotes the VAE reconstruction error, representing deviations from nominal latent manifold behavior;
    \item $I_t$ corresponds to the Isolation Forest score applied to standardized residuals for unsupervised outlier detection.
\end{itemize}

The coefficients $\alpha, \beta, \gamma, \delta$ control the relative weighting among the sub-components, and are selected empirically via grid-search to balance forecasting, reconstruction, and residual-based anomaly detection within the hybrid meta-learning ensemble.

\section{Methodology}

The proposed framework integrates physics-based simulation with hybrid deep learning models and residual-based meta-learning for anomaly forecasting in nonlinear dynamical systems.

\subsection{Nonlinear Synthetic Simulator}

We model a linked nonlinear system with random perturbations that displays hybrid growth-relaxation dynamics. The evolution during the "excitation" phase ($t \le t_{\rm ramp}$), with a single representative system state variable $P$ as the focus, is controlled by:

\[
\frac{dP}{dt}
  = P_{\rm coeff}\,P\Bigl(1-\frac{P}{P_{\rm sat}}\Bigr)
  + \underbrace{\frac{T_{\rm coeff}\,V}{w\,\tau}}_{P_{\rm RF}}
  + \underbrace{\eta(t)}_{\substack{\text{random}\\\text{perturbation}}},
\]

as well as throughout the "relaxation" stage ($t > t_{\rm ramp}$):
The formula is,
\[
\frac{dP}{dt}
  = -\alpha\,\bigl(P - P_0\bigr).
\]

In this case, saturation levels based on system characteristics are represented by $P_{\rm sat}$ and $T_{\rm sat}$. When gradient thresholds are reached, stochastic disturbances are introduced by $(V, \tau, w)$, $\eta(t)$, and $\alpha$ indicates the relaxation rate back to baseline.

\subsection{DA-RNN Training}

DA-RNN uses dual-stage attention: input attention selects relevant driving series; temporal attention highlights informative time steps.

\begin{algorithm}[H]
\caption{DA‐RNN One‐Step Forecasting}
\begin{algorithmic}[1]
  \Require Window length \(T\), dataset \((X,Y)\)
  \Require Epochs \(E\), batch size \(B\)
  \State Initialize DA‐RNN model and optimizer
  \For{\(\mathrm{epoch}=1\) \textbf{to} \(E\)}
    \For{each batch \((X_b,Y_b)\) of size \(B\)}
      \State \(\hat Y_b\gets \mathrm{DA\mbox{-}RNN}(X_b)\)
      \State \(\mathrm{loss}\gets \mathrm{MSE}(\hat Y_b,Y_b)\)
      \State optimizer.zero\_grad()
      \State loss.backward()
      \State optimizer.step()
    \EndFor
  \EndFor
  \State \Return trained DA‐RNN
\end{algorithmic}
\end{algorithm}

\subsection{CNN-LSTM Forecasting Model}

CNN-LSTM applies convolutional filters for local feature extraction, followed by LSTM for sequential forecasting.

\begin{algorithm}[H]
\caption{CNN–LSTM Forecasting}
\begin{algorithmic}[1]
  \Require Dataset \((X,Y)\), window length \(T\)
  \Require Epochs \(E\), batch size \(B\)
  \State Initialize CNN–LSTM model and optimizer
  \For{\(\mathrm{epoch}=1\) \textbf{to} \(E\)}
    \For{each batch \((X_b,Y_b)\) of size \(B\)}
      \State \(\hat Y_b\gets \mathrm{CNN\mbox{-}LSTM}(X_b)\)
      \State \(\mathrm{loss}\gets \mathrm{MSE}(\hat Y_b,Y_b)\)
      \State optimizer.zero\_grad()
      \State loss.backward()
      \State optimizer.step()
    \EndFor
  \EndFor
  \State \Return trained CNN–LSTM
\end{algorithmic}
\end{algorithm}

\subsection{Variational Autoencoder (VAE)}

The VAE reconstructs nominal samples, and reconstruction error flags anomalies.

\begin{algorithm}[H]
\caption{VAE Anomaly Scoring}
\begin{algorithmic}[1]
  \Require Data \(X\), latent dim \(L\)
  \Require Epochs \(E\), batch size \(B\)
  \State Initialize VAE (encoder, decoder) and optimizer
  \For{\(\mathrm{epoch}=1\) \textbf{to} \(E\)}
    \For{each batch \(x_b\) of size \(B\)}
      \State \((\hat x_b,\mu,\log\sigma)\gets \mathrm{VAE}(x_b)\)
      \State \(\ell_{\rm recon}\gets \mathrm{MSE}(\hat x_b,x_b)\)
      \State \(\ell_{\rm KL}\gets \mathrm{KL}(\mu,\log\sigma)\)
      \State \(\mathrm{loss}\gets \ell_{\rm recon} + \ell_{\rm KL}\)
      \State optimizer.zero\_grad()
      \State loss.backward()
      \State optimizer.step()
    \EndFor
  \EndFor
  \State \Return trained VAE
\end{algorithmic}
\end{algorithm}

\subsection{IF Anomaly}

Isolation Forest isolates anomalies in the standardized model residuals.

\begin{algorithm}[H]
\caption{Train Isolation Forest on Residuals}
\begin{algorithmic}[1]
  \Require Residual sequence \(R=[r_1,\dots,r_N]\)
  \Require Contamination \(c\)
  \Ensure Isolation Forest model \texttt{if\_model}, Standard scaler \texttt{scaler}
  \State \texttt{scaler} \(\gets\) \texttt{StandardScaler().fit}(R)
  \State \(R_s\gets\) \texttt{scaler.transform}(R)
  \State \texttt{if\_model}\(\gets\)\texttt{IsolationForest(contamination=}c\texttt{)}
  \State \texttt{if\_model.fit}(\(R_s\))
  \State \Return \texttt{if\_model}, \texttt{scaler}
\end{algorithmic}
\end{algorithm}

\subsection{Anomaly Forecasting}

At each time \(t\), we collect the DA‐RNN and CNN‐LSTM one‐step predictions over the horizon together with the Isolation Forest decision scores into a feature vector \(\mathbf{v}_t\).  A meta‐model \(f_{\mathrm{meta}}\) then produces a composite anomaly score
\[
\mathrm{score}_t = f_{\mathrm{meta}}(\mathbf{v}_t).
\]
An anomaly is flagged at time \(t\) if
\[
\mathrm{score}_t > b
\quad\text{and}\quad
\mathrm{score}_t - \mathrm{score}_{t-1} > \delta,
\]
where \(b\) is a baseline threshold and \(\delta\) is the minimum sudden‐change threshold. 

\begin{algorithm}[H]
\caption{Anomaly Forecasting via Meta-Model Mixture}
\begin{algorithmic}[1]
  \Require Data \(X_{1:N},\,y_{1:N}\), look-back window \(T\), horizon \(H\)  
  \Require Trained models: DA-RNN \(f_{\mathrm{da}}\), CNN-LSTM \(f_{\mathrm{cnn}}\), Isolation Forest \(\mathrm{IF}\), Meta-model \(f_{\mathrm{meta}}\)  
  \Require Baseline threshold \(b\), change threshold \(\delta\)
  \State \(\mathrm{anomalies}\gets\varnothing,\quad \mathrm{prev\_score}\gets 0\)
  \For{\(t = T\) \textbf{to} \(N-H\)}
    \State \textbf{Initialize Arrays} \(\mathrm{da\_preds}[1{:}H],\;\mathrm{cnn\_preds}[1{:}H]\)
    \Statex \quad and \(\mathrm{iso\_scores}[1{:}H]\)
    \For{\(h = 1\) \textbf{to} \(H\)}
      \State \(\mathrm{da\_preds}[h]\gets f_{\mathrm{da}}(X_{t:t+T},h)\)
      \State \(\mathrm{cnn\_preds}[h]\gets f_{\mathrm{cnn}}(X_{t:t+T},h)\)
      \State \(r\gets \lvert y_{t+h} - \mathrm{da\_preds}[h]\rvert\)
      \State \(\mathrm{iso\_scores}[h]\gets \mathrm{IF.decision\_function}(r)\)
    \EndFor
    \State \(\mathrm{feature\_vec}\gets [\,\mathrm{da\_preds};\,\mathrm{cnn\_preds};\,\mathrm{iso\_scores}\,]\)
    \State \(\mathrm{score}\gets f_{\mathrm{meta}}(\mathrm{feature\_vec})\)
    \If{\(\mathrm{score} > b\) \textbf{and} \(\mathrm{score}-\mathrm{prev\_score}>\delta\)}
      \State append \(t\) to \(\mathrm{anomalies}\)
    \EndIf
    \State \(\mathrm{prev\_score}\gets \mathrm{score}\)
  \EndFor
  \State \Return \(\mathrm{anomalies}\)
\end{algorithmic}
\end{algorithm}

\section{Results and Discussion}

Main goal of the meta-learner is to detect anomalies in the forecasted future of the system and make control system act up-on state changes based on predictions of the model itself, rather thresholding based on raw measurements. Models are trained for undershoot and overshoot the local and general behaviors and at the end fused together to extract a generalized behavior prediction. Furthermore, anomalies via the end model are then identified when the actual system behavior diverges from these learned forecasts, as reflected by forecast residuals, attention shifts, reconstruction errors, and residual-based outlier scores. This forecast-driven anomaly detection allows for early identification of both gradual drifts and sudden perturbations, even in the absence of explicit fault models or labeled anomalies.

\subsection{Simulator Behavior}

\begin{figure}[htbp]
\centering
\begin{subfigure}{0.20\textwidth}
  \includegraphics[width=\linewidth]{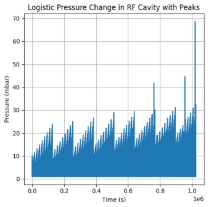}
  \caption{Simulated Main State Dynamics Evolution}
  \label{fig:pressure}
\end{subfigure}
\hfill
\begin{subfigure}{0.20\textwidth}
  \includegraphics[width=\linewidth]{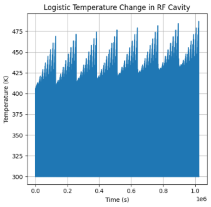}
  \caption{Simulated Auxiliary State Evolution}
  \label{fig:temperature}
\end{subfigure}
\caption{Simulated Nonlinear System State Dynamics: Simple RF system}
\label{fig:simulator}
\end{figure}

Fig.~\ref{fig:simulator} illustrates the synthetic system behavior generated by the nonlinear simulator. The logistic growth component drives smooth state increases, while injected random perturbations based on gradient-rules generate irregular transient peaks that simulate nonstationary system excursions. The secondary system variable exhibits smaller amplitude variations compared to the primary state, which is related to exogeneous parameters.

\subsection{Prediction Performance}

\begin{figure}[htbp]
\centering
\begin{subfigure}{0.31\textwidth}
  \includegraphics[width=\linewidth]{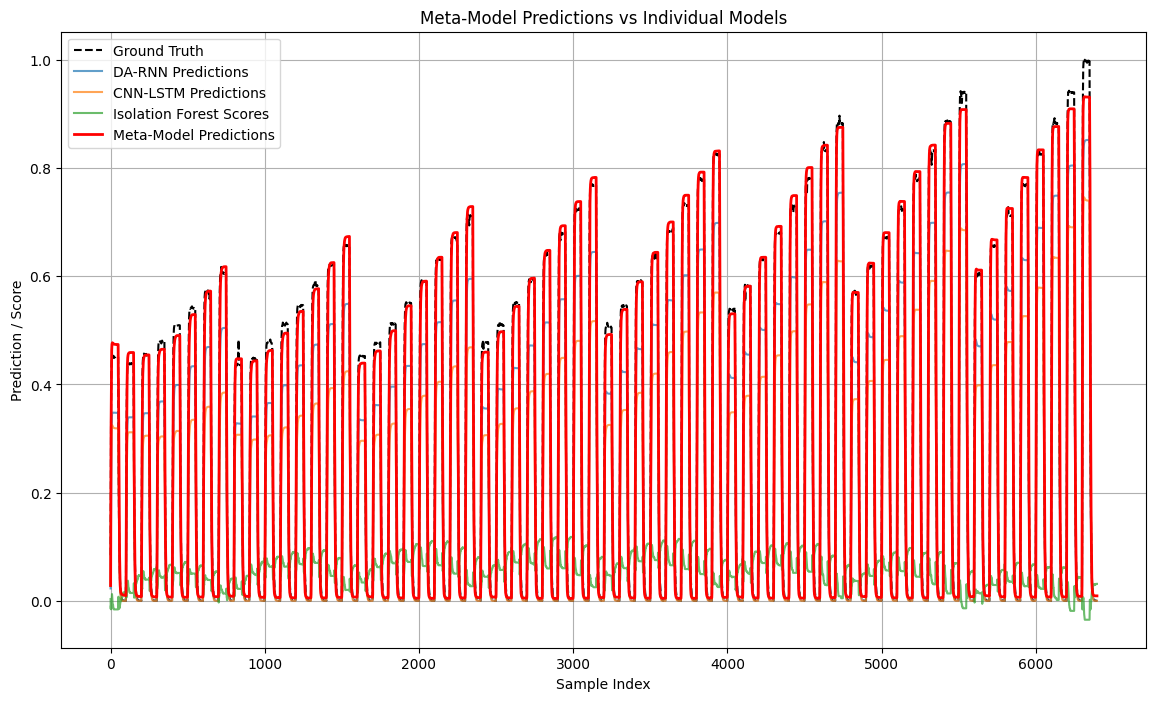}
  \caption{Model prediction comparison}
  \label{fig:prediction}
\end{subfigure}
\hfill
\begin{subfigure}{0.31\textwidth}
  \includegraphics[width=\linewidth]{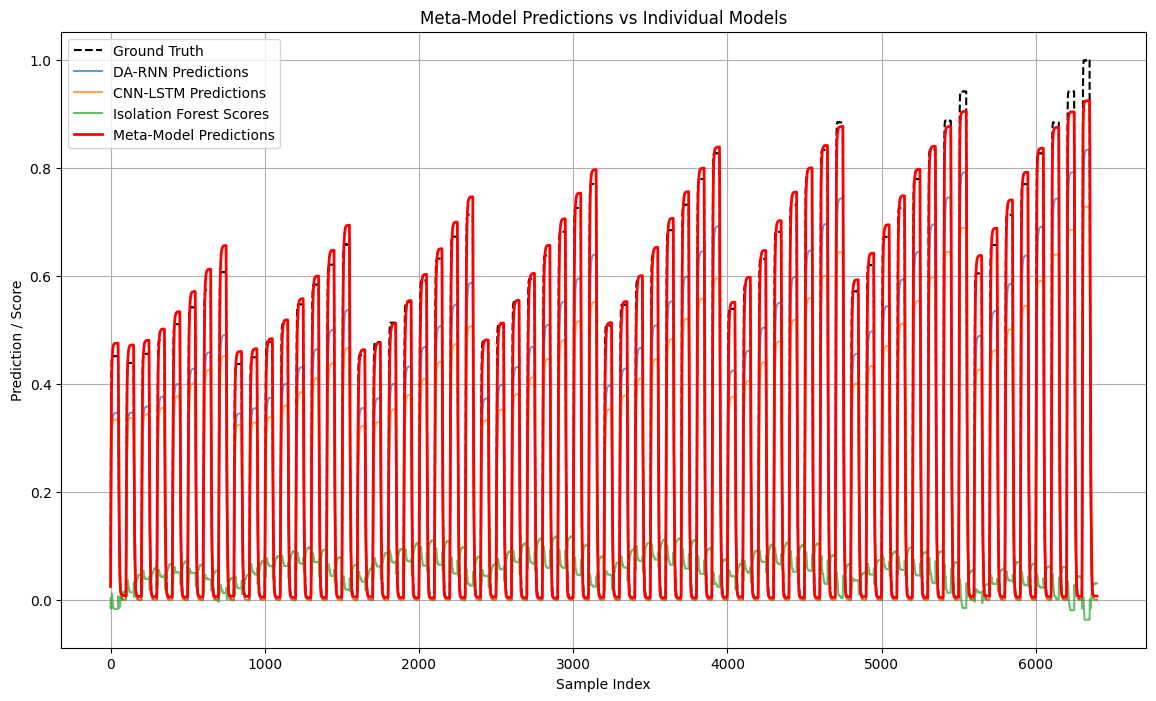}
  \caption{Initial residual error analysis}
  \label{fig:residual}
\end{subfigure}
\hfill
\begin{subfigure}{0.31\textwidth}
  \includegraphics[width=\linewidth]{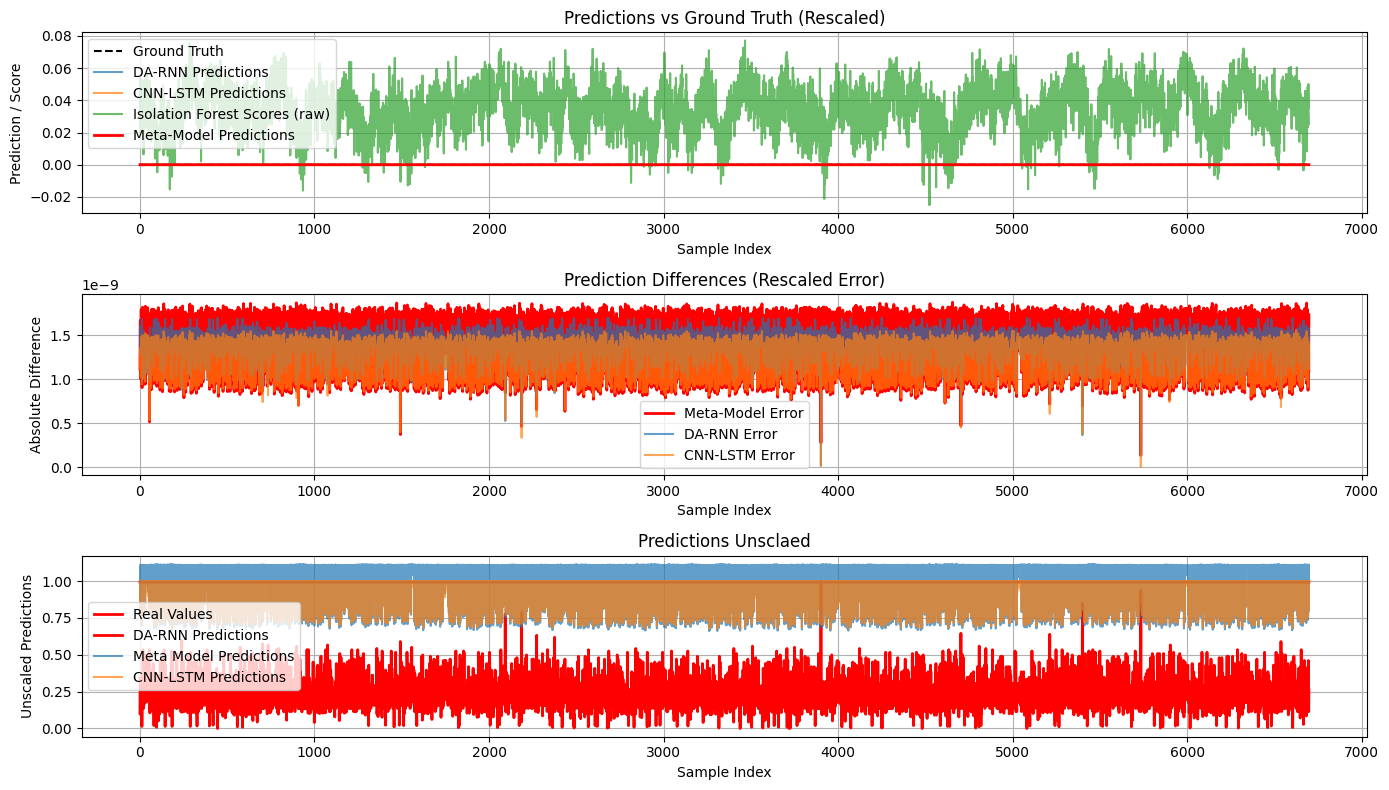}
  \caption{Sparse anomaly behavior}
  \label{fig:sparse}
\end{subfigure}
\caption{Forecasting and residual analysis for meta-model system}
\label{fig:predictions}
\end{figure}

Fig.~\ref{fig:predictions} compares forecasting accuracy across model components:

\begin{itemize}
    \item The individual forecasting models (DA-RNN, CNN-LSTM) track standard system behavior under regular conditions.
    \item The hybrid meta-learning ensemble improves deviation capture during nonlinear regime changes and transient events.
    \item Isolation Forest applied on residuals enhances the detection of unpredictable perturbations not directly captured by the forecasting models.
\end{itemize}

\subsection{Anomaly Localization}

\begin{figure}[htbp]
\centering
\includegraphics[width=0.45\textwidth]{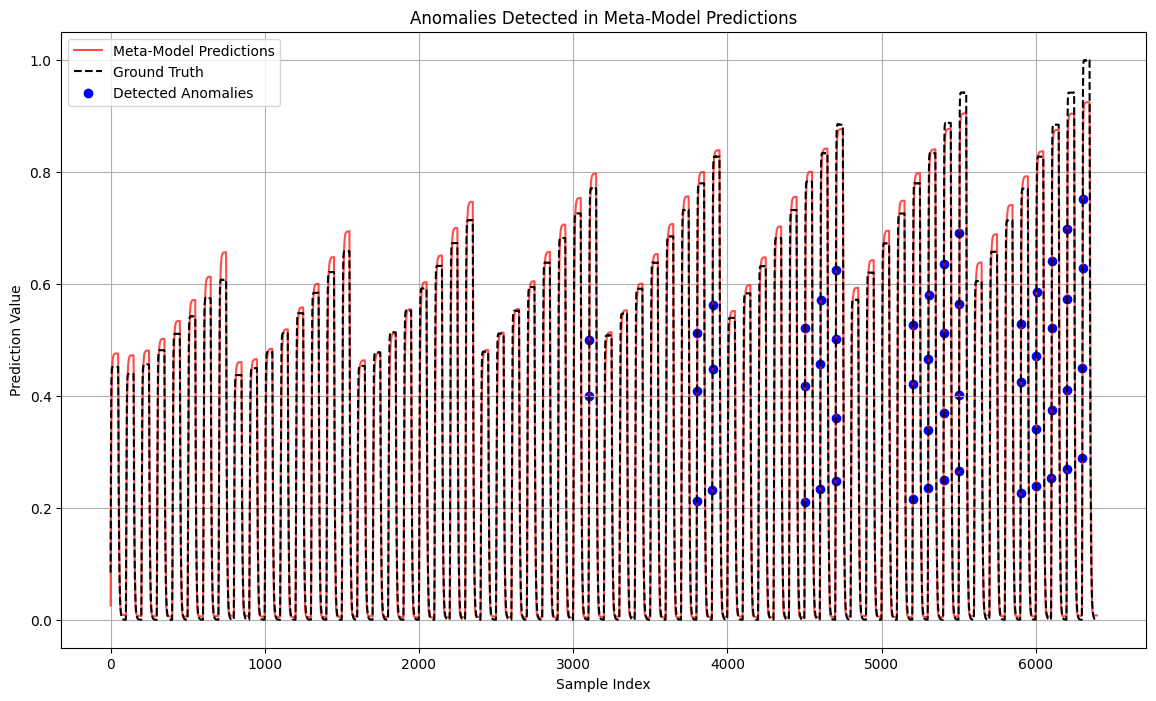}
\caption{Anomaly detection results: detected anomalies (blue markers) overlaid on simulation signal, over a threshold hyperparameter}
\label{fig:anomalies}
\end{figure}

As shown in Fig.~\ref{fig:anomalies}, detected anomalies (blue marker-dots) correspond closely with injected perturbations in the simulation. The meta-learning ensemble effectively localizes abrupt deviations while maintaining stability.

\subsection{Error Analysis}

Residual analysis (Fig.~\ref{fig:predictions}) demonstrates that the meta-model ensemble achieves lower absolute forecast error compared to standalone forecasting models. The inclusion of Isolation Forest improves robustness by capturing unexpected variance in residual patterns.

\subsection{Attention Visualization}

\begin{figure}[htbp]
\centering
\includegraphics[width=0.95\linewidth]{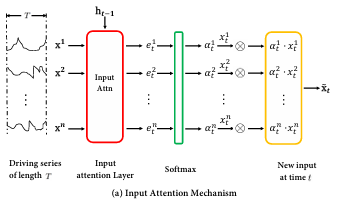}
\caption{Input attention network \cite{qin2017dual}}
\label{fig:inputattention}
\end{figure}

\begin{figure}[htbp]
\centering
\includegraphics[width=\linewidth]{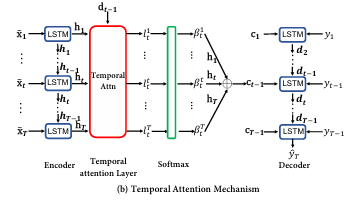}
\caption{Temporal attention network \cite{qin2017dual}}
\label{fig:temporalattention}
\end{figure}

Figs.~\ref{fig:inputattention} and \ref{fig:temporalattention} visualize the learned attention mechanisms:

\begin{itemize}
    \item Input attention highlights dominant driving features contributing to system evolution.
    \item Temporal attention dynamically adapts to emphasize precursor signals, facilitating early anomaly detection.
\end{itemize}

\subsection{Predictive Evaluation of State Changes}

Beyond automated and passive monitoring, the forecast-driven framework enables predictive evaluation of proposed control actions. By simulating candidate state transitions through the trained forecasting models, the system can estimate potential anomaly scores before applying control changes. This allows preemptive rejection of unsafe actions that may drive the system into fault regions, enhancing operational safety and autonomous decision-making.



\section{Conclusion}

We have introduced a novel hybrid meta‐learning strategy for anticipating anomalies in complex nonlinear dynamical systems by combining physics-inspired simulations with an ensemble of deep learning models. Rather than applying simple thresholds to raw measurements, our method predicts future system trajectories (via DA-RNN and CNN-LSTM), reconstructs likely system states (using a VAE), and evaluates residual deviations (with an Isolation Forest). A lightweight meta-learner then fuses these distinct anomaly indicators into a unified score, yielding early warnings for both slow drifts and sudden disruptions.

Importantly, by forecasting candidate future states before control commands are issued, this framework enables proactive screening of potentially unsafe actions—extending its utility from passive monitoring to active decision support in safety-critical environments. The design is fully data-driven, agnostic to specific physical models, and readily transferable to diverse nonlinear systems even when detailed equations or large labeled datasets are lacking.

Looking forward, integrating sparse system identification techniques (SINDy) could further enrich the approach by injecting partial physics knowledge. This modularity offers a clear path for adapting the framework to scenarios with varying levels of prior information, thereby broadening its applicability across engineering and scientific domains.

\section{Future Work}

Future directions include several extensions to improve both the generality and deployment-readiness of the proposed framework:

\begin{itemize}
    \item Validate the methodology on real-world nonlinear system datasets from industrial, cyber-physical, or process control domains.
    \item Explore transfer learning adaptations to hardware platforms for real-time deployment using embedded controllers and high-level synthesis (HLS) for online protection and self-tuning.
    \item Extend the meta-learning framework to dynamically adjust anomaly thresholds based on drift-adaptive baselines and long-term system degradation.
    \item Incorporate explainability and physics-informed surrogate models to improve interpretability and operator trust.
    \item Integrate system identification methods such as Sparse Identification of Nonlinear Dynamical Systems (SINDy) to extract governing equations from partial measurements and enable model-informed training pipelines when physical knowledge is limited.
\end{itemize}

\bibliographystyle{IEEEtran}
\bibliography{refs}

\end{document}